\documentclass[conference]{IEEEtran}
\IEEEoverridecommandlockouts

\usepackage{booktabs} 

\usepackage{cite}
\usepackage{amsmath,amssymb,amsfonts}
\usepackage{algorithmic}
\usepackage{graphicx}
\usepackage{textcomp}
\usepackage{xcolor}
\def\BibTeX{{\rm B\kern-.05em{\sc i\kern-.025em b}\kern-.08em
    T\kern-.1667em\lower.7ex\hbox{E}\kern-.125emX}}

\makeatletter
\newcommand{\linebreakand}{%
  \end{@IEEEauthorhalign}
  \hfill\mbox{}\par
  \mbox{}\hfill\begin{@IEEEauthorhalign}
}
\makeatother

\begin{document}

\title{MPLite: Multi-Aspect Pretraining for Mining Clinical Health Records}

\author{
\IEEEauthorblockN{Eric Yang\IEEEauthorrefmark{2}}
\IEEEauthorblockA{\textit{Department of Computer Science} \\
\textit{Stevens Institute of Technology}\\
Hoboken, NJ,  USA \\
eyang6@stevens.edu}
\and
\IEEEauthorblockN{Pengfei Hu\IEEEauthorrefmark{2}}
\IEEEauthorblockA{\textit{Department of Computer Science} \\
\textit{Stevens Institute of Technology}\\
Hoboken, NJ,  USA \\
phu9@stevens.edu}
\linebreakand
\IEEEauthorblockN{Xiaoxue Han}
\IEEEauthorblockA{\textit{Department of Computer Science} \\
\textit{Stevens Institute of Technology}\\
Hoboken, NJ, USA \\
xhan26@stevens.edu}
\and
\IEEEauthorblockN{Yue Ning\IEEEauthorrefmark{1}}
\IEEEauthorblockA{\textit{Department of Computer Science} \\
\textit{Stevens Institute of Technology}\\
Hoboken, NJ, USA \\
yue.ning@stevens.edu}
}

\IEEEoverridecommandlockouts
\IEEEaftertitletext{\vspace{-1.0\baselineskip}\noindent\thanks{\IEEEauthorrefmark{2} Equal contribution as co-first authors.}
\thanks{\IEEEauthorrefmark{1} Corresponding author}}

\maketitle

\newcommand{\system}{\textbf{MPLite}}

\begin{abstract}
The adoption of digital systems in healthcare has resulted in the accumulation of vast electronic health records (EHRs), offering valuable data for machine learning methods to predict patient health outcomes. 
However, single-visit records of patients are often neglected in the training process due to the lack of annotations of next-visit information, thereby limiting the predictive and expressive power of machine learning models. 
In this paper, we present a novel framework \system{} that utilizes Multi-aspect Pretraining with Lab results through a light-weight neural network to enhance medical concept representation and predict future health outcomes of individuals.
By incorporating both structured medical data and additional information from lab results, our approach fully leverages patient admission records. 
We design a pretraining module that predicts medical codes based on lab results, ensuring robust prediction by fusing multiple aspects of features. 
Our experimental evaluation using both MIMIC-III and MIMIC-IV datasets demonstrates improvements over existing models in diagnosis prediction and heart failure prediction tasks, achieving a higher weighted-F\textsubscript{1} and recall with MPLite.
This work reveals the potential of integrating diverse aspects of data to advance predictive modeling in healthcare.
\end{abstract}


\section{Introduction}
\begin{figure}[t]
    \centering
    \includegraphics[width=8cm]{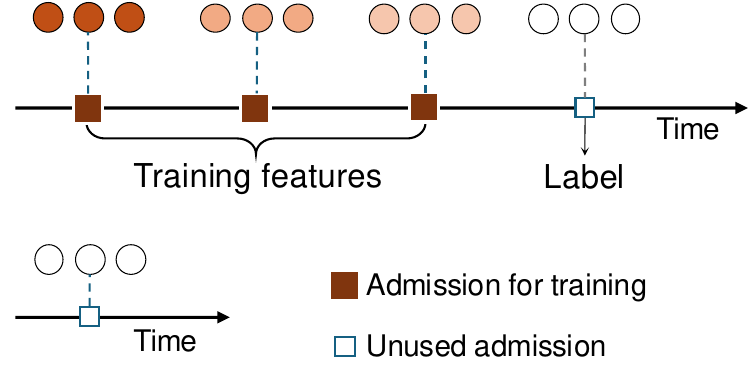}
    \caption{Example of traditional setting on most predictive model}
    \label{fig:single_admission}
\end{figure}

EHR datasets, such as MIMIC-III~\cite{mimic3db}, provide comprehensive medical information, including vital signs, diagnoses, medications, and lab results. These multi-aspect features are valuable resources for predicting personalized health events, such as diagnosis predictions. 
Meanwhile, deep learning technique have become a common approach for analyzing sequential data within healthcare~\cite{ChoiBSSS16, zhang2023event, zhang2023text}. However, many studies often exclude patient examples with only single-visit records, since these records lack labels for prediction tasks involving future admissions.
For instance, when training a supervised machine learning model to predict diagnoses in the next visit given previous visits in the MIMIC-III dataset, we need the annotations/labels for the next visit. 
Therefore, temporal prediction models rely on patient data with at least two visits to complete the training process.
Single-visit records are not fully utilized in training predictive models as shown in Figure~\ref{fig:single_admission}.
However, multi-visit patients contribute to only a small portion of the dataset. Among a total of 46,520 patients, only 16.20\% have multiple visits. The remaining 83.80\% are single-visit patients, which could also provide rich information for models to learn useful patterns and make better predictions.

To fully utilize these single admission records in an EHR dataset, there are two popular solutions to address this issue:
(1) Transformer models like G-BERT~\cite{ShangMXS19} leverage single-admission data to design customized self-supervised tasks, which typically treat medical concepts or admissions as masked tokens and further enhance intermediate representation learning within the encoder framework. 
(2) Multi-aspect learning~\cite{LuRCKN21, ChoiXLDFXD20} incorporates diverse features, such as lab test results or clinical notes, to enrich the representation learning of medical concepts, which helps models better capture the complexity and interrelationships inherent in medical data.
The former approach, although widely adopted by early studies~\cite{Rasmy0XTZ21, poulain2024graph}, is susceptible to the order of medical codes and may not be lightweight enough to function as a plug-and-play module. In contrast, while the latter approach demands high-quality and relevant additional medical concepts, it enables models to learn collaborative representations, leading to more accurate predictions with the addition of a lightweight module.

In this study, we leverage single admissions as auxiliary training data to predict diagnoses and health risks, such as heart failure. 
Recognizing the pivotal role that lab test results play prior to training, we propose a novel framework, \system{}, which is an additional plug-in-and-play module to learn relationships between lab results and diagnoses through a \textbf{M}ulti-aspect \textbf{P}retraining and ``\textbf{Lite}'' module.
This framework captures the underlying patterns and associations that are present in both multiple-visit and single-visit data. 
We then illustrate how incorporating this pre-trained knowledge can significantly enhance the predictive capabilities of temporal neural networks, particularly for forecasting health risks in patients with multiple visits.
By fine-tuning the pre-trained subnetwork on two specific health risk prediction tasks, we demonstrate the effective extraction of valuable insights from abundant single-visit patient data. 
The pretraining module underscore the advantages of pretraining on diverse medical features beyond diagnoses concepts and highlights the broader applicability of lab test data in predictive healthcare.

\section{Related Work}
\label{sec:related_work}
Deep learning models have been extensively applied to electronic health records (EHR) to extract representations of medical patterns, addressing various real-world healthcare prediction tasks like diagnosis prediction.

\subsection{CNN/RNN-based Models}
Most early studies in this area can be categorized into two main subcategories:
(i) RNN-based models, where predictive methods like GRU~\cite{ChoMGBBSB14}, RETAIN~\cite{ChoiBSKSS16}, and Timeline~\cite{BaiZEV18} combine attention mechanisms and RNN for prediction. Other models \cite{ChoiBSSS16, MaCZYSG17, MaZWRWTMGG20} leverage RNNs to handle time-series data effectively; 
(ii) CNN-based models, such as Deepr \cite{NguyenTWV17} and AdaCare \cite{MaGWZWRTGM20}, use convolution and pooling layers to process features in EHR. 
However, these methods often overlook relations among encoded medical concepts and other critical aspects such as lab test results.

\subsection{Graph/Transformer-based Models}
Recently, there has been a trend towards using ontology graphs to incorporate additional information related to medical concepts in predictions such as GRAM\cite{ChoiBSSS17}, G-Bert~\cite{ShangMXS19}, GCT~\cite{ChoiXLDFXD20}, Variationally Regularized GNN~\cite{ZhuR21}, GraphCare~\cite{jiang2023graphcare}, ME2Vec~\cite{wu2021leveraging}, RGNN~\cite{LiuLDTWCYZ20}. 
However, most existing works primarily rely on admission medical concepts as features for various deep learning models. 
Meanwhile, following the success of the transformer architecture, researchers have quickly adopted it for EHR data. Encoder-decoder structures offer the advantage of fully utilizing single-visit data in the pretraining process by customizing proxy tasks for different prediction tasks. 
Early studies, like G-Bert~\cite{ShangMXS19} treat medical codes as tokens and incorporate hierarchical domain knowledge along with diagnosis codes.
Recent models like HiTANet~\cite{LuoYXM20}, Med-BERT~\cite{Rasmy0XTZ21}, and Sherbet~\cite{LuRN23} have also been trained to precisely identify patient information based on various medical concepts.
However, the pretraining phase in most works cannot be easily separated into a plug-and-play module, limiting its generalizability when transferring pretraining information to new tasks or different structures.

\subsection{Models with Mutli-Aspect Features}
Beyond traditional medical concepts, such as condition, medication, and treatment codes, researchers~\cite{poulain2024graph} also involve additional information (e.g. demographic features and timestamps) in each admission record.
To augment representation from different modalities, both CGL~\cite{LuRCKN21} and MedGTX~\cite{ParkBKKC22} integrates disease-patient graphs and unstructured text from clinical notes through encoder structures to demostrates the importance of involving additional information other than sequence of medical concepts. 
MiME~\cite{ChoiXSS18} and GCT~\cite{ChoiXLDFXD20} are preliminary tries to involve lab results as input features to further optimize medical hidden representation.
However, these integrated models cannot work well in the absence of corresponding records, and they always have complicated preprocessing or fusion steps which cannot be generalized as lightweight modules.

In this paper, we propose \system{} that allows different models to jointly learn representations of medical diagnosis codes and lab results. Our framework provides a novel perspective for integrating different features to achieve more accurate predictions. The experimental results demonstrate a significant improvement with the extensive pretraining module in predicting health outcomes over several baselines, as confirmed by confidence intervals obtained from repeated experiments.

\section{Proposed Method}
\label{sec:method}

We begin by describing the notations and then introduce our proposed framework, which includes a pretraining module with lab results, along with instructions on how to seamlessly integrate and utilize the module for downstream tasks.

\begin{table}
\small
\centering
 \caption{Notations Used in This Paper}
 \vspace{-1em}
        \begin{tabular}{cp{6cm}} 
        \toprule
        \textbf{Notation} & \textbf{Definition}\\ [0.5ex] 
        \midrule
        $\mathcal{S}$ & EHR dataset \\
        $P_i$ & $i$-th patient \\
        $\mathcal{C},\mathcal{D},\mathcal{L}$ & Sets of medical concepts, diagnosis codes, and lab test codes\\
        $|\mathcal{C}|,|\mathcal{D}|,|\mathcal{L}|$ & Cardinality of medical concepts, diagnoses, and lab test codes\\
        $T_{i}$ & The number of visits for patient $\mathbf{p_i}$ \\ 
        $\mathbf{x}_t,\mathbf{x}_{t}^{D},\mathbf{x}_{t}^{L}$ & 
        Multi-hot vector for the $t$-th visit of a patient \\
        \bottomrule
        \end{tabular}
        \vspace{-1em}
\end{table}

\begin{figure*}[ht]
    \centering
    \includegraphics[width=14cm]{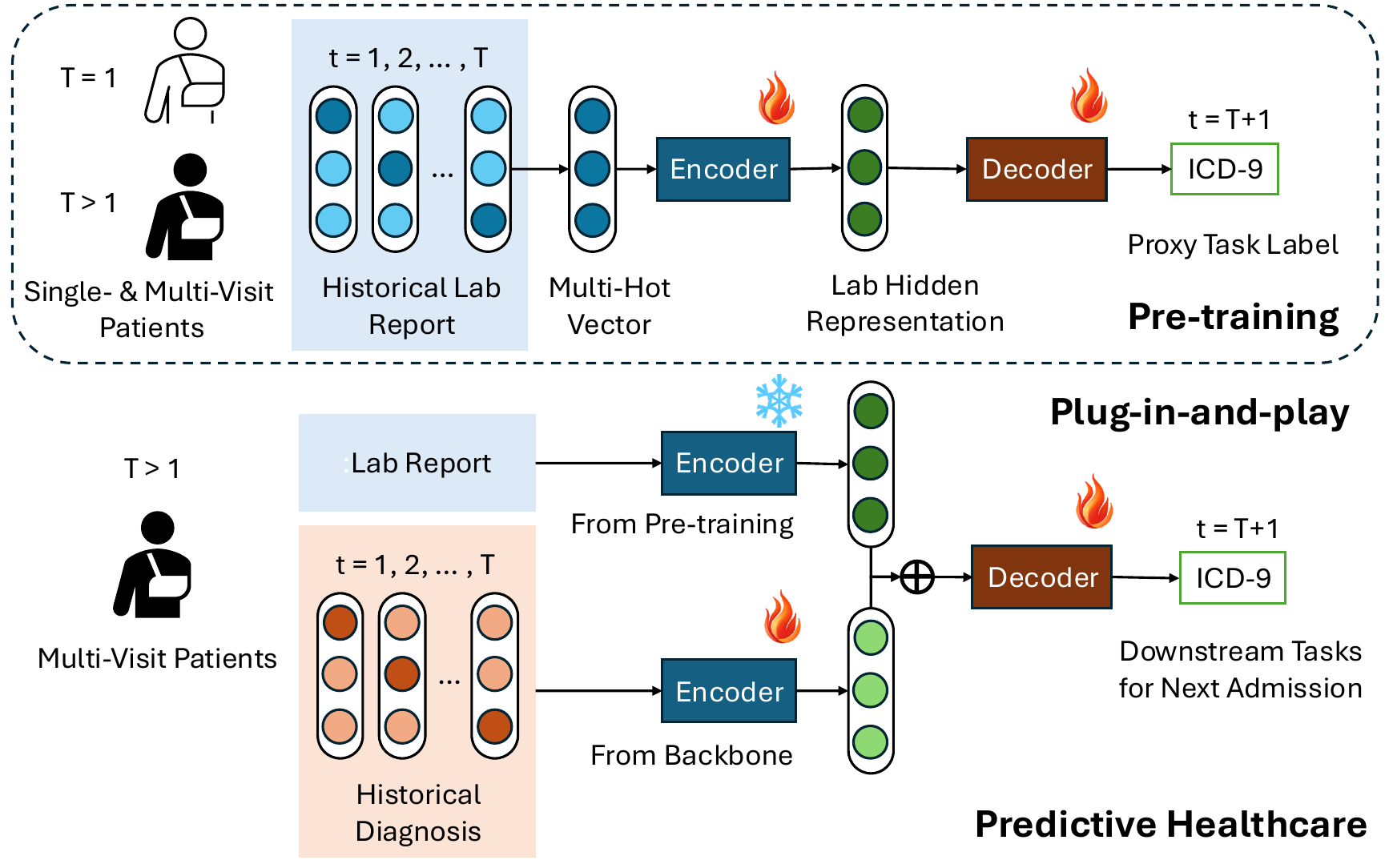}
    \caption{Overview of the proposed MPLite Framework}
    \label{fig:pretrained}
\end{figure*}

\subsection{General Notations}
\label{subsec:notation}
An EHR dataset $\mathcal{S}$ is a collection of patient admission records of N patients $\{P_1,P_2,...,P_N\} \in \mathcal{S}$ in total.
For admission records of each patient, the \(i\)-th patient can be represented as a sequence of $T_i$ admission records 
$\{\mathbf{x}_1, \mathbf{x}_2,.., \mathbf{x}_{T_i}\} \in \mathbf{p}_i$ in chronological order, where $T_i$ is the number of admissions for the patient.
The goal of our predefined prediction tasks is to predict the label at the end of the sequence, \(\mathbf{y} \in \{0, 1\}^s\), which can be either a one-hot or multi-hot vector.
We then omit $i$ in the rest of the sections and explain our framework using single patient to avoid misunderstanding.

Specifically, a single admission $\mathbf{x}_t$ ($t\in\{1\}$) can be also represented as a multi-hot vector with dimensions corresponding to the medical concepts \(\mathcal{C}=\{c_1, c_2,..., c_{|\mathcal{C}|}\}\) where $|\mathcal{C}|$ is the total number of medical concepts. Each element in the vector is a boolean value indicating the presence (1) or absence (0) of the corresponding medical concept.
Note that, we consider both \textit{ITEM\_ID} from lab results and \textit{ICD-9} codes from diagnoses as medical concepts in our experiment, thus medical concepts might be either diagnosis codes $\mathbf{x}^{D}_t$ within vocabulary $\mathcal{D}\in\mathcal{C}$ or lab items codes $\mathbf{x}^{L}_t$ within vocabulary $\mathcal{L}\in\mathcal{C}$ in single admission.
As medical concepts depend on problem formulation and real-world EHR data, procedures, drugs, and some other medical concepts can also be considered medical codes from a broader perspective.
In the following sections, we also use abstract symbols like \textit{MLP} to denote specific frameworks with mutable settings.

\subsection{MPLite Framework}
\subsubsection{Multi-Aspect Pretraining}
To fully leverage the EHR data, it is essential to utilize records from single-admission patients, who constitute the majority of the dataset.
Since these records lack labels for future admissions, our focus is on learning the relationship between lab test results and diagnoses from the current visit.
A single-visit patient has only one admission record, so we consider a single-visit patient equivalent to a single visit in this part.
We hypothesize that additional aspects of features (e.g, lab tests) reflect important information about a patient's existing diagnoses.
Thus, we identify lab results as additional medical concepts for each patient, considering that lab results are one of the most crucial components in describing diagnoses results.

In terms of the pretraining step of Figure~\ref{fig:pretrained}, we define a novel proxy task that predicts the diagnoses shown in the sequence of visits by historical lab results in the pretraining step.
As mentioned in the section \ref{subsec:notation}, diagnoses and lab results sets are denoted by $\mathcal{D}$ and $\mathcal{L}$ respectively, and we aim to decode the item code set from lab results $\mathbf{x}^{L}$ into the probability distribution 
\(\hat{\mathbf{y}} = P(\mathbf{x}^{D}|\mathbf{x}^{L})\) for each patient.
Here we use a multi-layer perceptron (MLP) for parametrization to transform lab results to diagnoses of patients:
\begin{align}
 \hat{\mathbf{y}} &= \sigma (\text{MLP}(\mathbf{x}^{L}_{1} \mid P_{\text{single}})) \\
 \hat{\mathbf{y}} &= \sigma (\text{MLP}(\text{Integrate}(\{\mathbf{x}^L_t\}_{t=1}^T) \mid P_{\text{multi}}))
\end{align}
Here $P_{\text{single}},P_{\text{multi}}$ denotes single-visit and multi-visit patients, and $\sigma$ means the activation function.
There are two main reasons we chose MLP as the backbone model for the pretraining module:
(1) It directly predicts the probability distribution of diagnosis codes efficiently, requiring minimal computational resources.
(2) It achieves competitive predictive performance for the defined proxy task, even when compared to models incorporating embedding or convolution modules.
Moreover, \texttt{Integrate} means we integrate the sequence of multiple visit into a single vector, which can be aligned with the input of single-visit patients as shown in equation~\ref{eq:integrate}.
\begin{equation}
    \text{Integrate}(\{\mathbf{x}^L_t\}_{t=1}^T) = \bigvee_{t=1}^{T} \mathbf{x}^{L}_{t}
    \label{eq:integrate}
\end{equation}
For the lab result data, we assume that lab results are all up-to-date, and we considered single lab-test code $c^{(\mathcal{L})}_k$ normal if it has not been taken or was tested normal in the most recent test. 
For each patient, given lab results prior to the ($t+1$)-th visit is \(\mathbf{x}^{L}_{t}\in\mathbb{R}^{|\mathcal{L}|}\) ($|\mathcal{L}|=697$ in MIMIC-III dataset).
The defined proxy task is a multi-label classification task. 
Given multi-hot vector of lab results $\mathbf{x}^{L}_{t}\in\{0, 1\}^{|\mathcal{L}|}$, we use the first dense layer as encoder to get $H$-dimensional hidden representation $\mathbf{h}^{L}$ for each patient, and then we leverage the second dense layer as decoder to convert such hidden representation as diagnose classifier with output $\hat{y}$.
The pretraining dense layers and corresponding loss function are defined as follows:
\begin{align}
\mathbf{h}^{L} &= \text{Encoder}(\{\mathbf{x}^L_t\}_{t=1}^T \mid P)\in\mathbb{R}^{h}\\
\hat{\mathbf{y}} &= \text{Decoder}(\mathbf{h}^{L}) = \sigma(\textbf{w}_k\mathbf{h}^{L})\in\mathbb{R}^{|\mathcal{D}|}\\
\mathcal{L_{\text{patient}}} &= - \left[ \mathbf{y} \log(\hat{\mathbf{y}}) + (1 - \mathbf{y}) \log(1 - \hat{\mathbf{y}}) \right]
\end{align}

At the pretraining step, $t$ depends on the number of available admission records for each patient, and we use binary cross entropy as loss function through $N$ single-visit patients.
Note that, such module is learnable within both single-visit and multi-visit patients, and involved parameters are fixed after the pretraining process.
As a self-supervised learning problem, this proxy task is not simply input reconstruction and will not be affected by the order of medical concepts in a single visit. This is also the main advantage compared to traditional transformer-base models.
The pretraining process does not have access to the validation and test sets of the prediction model. Thus there are no data leakage issues.


\subsubsection{Integration and Inference}
Now let us focus on how to fuse both representations from a backbone prediction model and the proposed pretraining module.
Note that, subscript $t$ might be also involved in model in terms of the training setting across different baseline.
For example, some works feed model by admission-level data, which means patient with multiple visits can be fed consecutively into the model.
For the adaptation ability of our framework, we also transfer this setting into the description of our framework.

Assuming we already have the final output $\mathbf{o}_{t}\in\mathbb{R}^{|\mathcal{C}|}$ for prediction of the $t$-th admission before feeding into the classifier of existing baselines, we can also retrieve lab results vector $\mathbf{x}_{t}^{L}$ as the input of the pre-trained module.
$|\mathcal{C}|$ is the output dimension, which is also considered as the vocabulary size.
We keep the same format of input for $\mathbf{x}_{t}^{L}$ and get hidden representation $\mathbf{h}_{t}^{L}\in\mathbb{R}^{h}$ in terms of patient's lab results through the pretrained encoder dense layer.
We then use a classifier with single dense layer to get prediction $\hat{y}_t$ for multiple prediction tasks after concatenating both patient-level representations as shown in Figure~\ref{fig:pretrained}.
Finally, the integration step and classifier are defined as follows, the output dimension of classifier can be modified for various prediction tasks:
\begin{align}
\mathbf{o}_{t} &= \text{Encoder}(\{\mathbf{x}^L_t\}_{t=1}^T \mid P_{\text{multi}})\\
\mathbf{o}_{t}' &= \mathbf{o}_{t} \mathbin\Vert \mathbf{h}_{t}^{L} \in \mathbb{R}^{|\mathcal{C}|+h}\\
\hat{\mathbf{y}}_{t} &= \text{Classifier}(\mathbf{o}_{t}') \in \mathbb{R}^{|\mathcal{C}|}
\end{align}

We can still remain the same loss function $\mathcal{L}$ as the one already defined in the backbone module.
Through the definition of inference part, we can easily plug in pretraining module and optimize current model's output by integrating lab results for more precise prediction.

\subsection{Downstream Tasks}
The proposed framework can be adapted for various prediction tasks. Consider a patient with $T+1$ admission records, we can build one sample with admission history $\{\mathbf{x}_1,\mathbf{x}_2,...,\mathbf{x}_T\}$ for each patient.
We perform two prediction tasks in our experiments by the following definition:

(1) \textbf{Diagnosis (DG) Prediction} predicts the diagnosis result of the next admission given previous admission records. Formally, we learn a function
$f:(\mathbf{x}_1,
\mathbf{x}_2,...,
\mathbf{x}_t) \rightarrow \mathbf{y}[\mathbf{x}_{t+1}]$ where $t \leq T$ and
$\mathbf{y}[\mathbf{x}_{t+1}] \in \mathbb{R}^{|\mathcal{D}|}$ is a multi-hot vector where $|\mathcal{D}|$ denotes the number of all diagnosis codes.

(2) \textbf{Heart Failure (HF) Prediction} predicts if heart failure (i.e., ICD-9 prefixed code of 428) is diagnosed in the next admission. Formally, we learn a function
$f:(\mathbf{x}_1,\mathbf{x}_2,...,\mathbf{x}_t) \rightarrow \mathbf{y}[\mathbf{x}_{t+1}]$ where $t \leq T$ and
$\mathbf{y}[\mathbf{x}_{t+1}] \in \{0,1\}$ is a binary label indicating whether heart failure is diagnosed in the admission.

The binary cross-entropy (BCE) loss is used with a sigmoid function to train the learning framework for both binary and multi-label classifications tasks.

\begin{table}[t]
    \centering
    \small
        \caption{Statistics of the MIMIC-III dataset}
        \vspace{-1em}
    \label{tab:statistics}
    \begin{tabular}{@{}l r@{}} 
        \toprule
        \# patients in total & 46,520 \\
        \# patients with multiple visits & 7,537 \\
        \# patients with multiple visits utilized in experiments & 7,493 \\
        \# patients with single visit & 38,983 \\
        \# patients with single visit utilized in experiments & 26,085 \\
        Avg. visits per patient in MIMIC-III & 1.27 \\
        \midrule
        \# Medical codes (disease) & 4,880 \\
        \# Items (lab results) & 697 \\
        \bottomrule
    \end{tabular}
\end{table}

\section{Evaluation}
\label{sec:eval}

\setlength{\tabcolsep}{2pt}
\begin{table*}[t]
    \centering
    \small
    \caption{Prediction Results on MIMIC-III and MIMIC-IV for Diagnosis and Heart Failure Prediction. We report the average performance (\texttt{\%}) and standard deviation (in brackets) of each model over 10 runs. ``No'' in the Pretrain column means the original baselines, and ``+MPLite'' means that we plug in the pretraining module into the corresponding baselines}
    \label{tab:mimic_combined}
    \begin{tabular}{@{}lc|ccccc|ccccc@{}}
        \toprule
         & & \multicolumn{5}{c|}{\textbf{MIMIC-III}} & \multicolumn{5}{c}{\textbf{MIMIC-IV}} \\
         & & \multicolumn{3}{c}{\textbf{DG Prediction}} & \multicolumn{2}{c|}{\textbf{HF Prediction}} & \multicolumn{3}{c}{\textbf{DG Prediction}} & \multicolumn{2}{c}{\textbf{HF Prediction}}\\
        \textbf{Models} & \textbf{Pretrain} & \textbf{w-}$\mathbf{F_1}$ & \textbf{R@10} & \textbf{R@20} & \textbf{AUC} & $\mathbf{F_1}$ & \textbf{w-}$\mathbf{F_1}$ & \textbf{R@10} & \textbf{R@20} & \textbf{AUC} & $\mathbf{F_1}$ \\
        \midrule
        GRU & No & 17.82$_{(0.43)}$ & 31.56$_{(0.40)}$ & 33.64$_{(0.38)}$ & 80.54$_{(0.60)}$ & 68.93$_{(0.53)}$ & 19.55$_{(0.48)}$ & 35.12$_{(0.57)}$ & 37.91$_{(0.54)}$ & 81.33$_{(0.71)}$ & 69.31$_{(0.56)}$ \\
        GRU & \textsubscript{+MPLite} & 19.58$_{(0.34)}$ & 33.82$_{(0.39)}$ & 35.97$_{(0.35)}$ & 82.01$_{(0.55)}$ & 70.56$_{(0.47)}$ & 21.87$_{(0.37)}$ & 37.84$_{(0.43)}$ & 40.63$_{(0.48)}$ & 83.12$_{(0.62)}$ & 71.02$_{(0.42)}$ \\
        \midrule
        Dipole & No & 14.66$_{(0.21)}$ & 28.73$_{(0.28)}$ & 29.44$_{(0.20)}$ & 82.08$_{(0.45)}$ & 70.35$_{(0.51)}$ & 17.16$_{(0.36)}$ & 32.21$_{(0.30)}$ & 38.74$_{(0.32)}$ & 84.80$_{(0.47)}$ & 69.52$_{(0.44)}$ \\
        Dipole & \textsubscript{+MPLite} & 18.27$_{(0.30)}$ & 30.91$_{(0.37)}$ & 32.97$_{(0.29)}$ & 83.56$_{(0.53)}$ & 71.53$_{(0.46)}$ & 20.63$_{(0.33)}$ & 38.12$_{(0.36)}$ & 40.75$_{(0.41)}$ & 85.67$_{(0.56)}$ & 71.02$_{(0.50)}$ \\
        \midrule
        Deepr & No & 11.68$_{(0.17)}$ & 26.47$_{(0.15)}$ & 27.53$_{(0.12)}$ & 81.36$_{(0.39)}$ & 69.54$_{(0.49)}$ & 18.58$_{(0.31)}$ & 36.79$_{(0.29)}$ & 39.45$_{(0.21)}$ & 83.61$_{(0.50)}$ & 70.46$_{(0.53)}$ \\
        Deepr & \textsubscript{+MPLite} & 18.43$_{(0.28)}$ & 31.08$_{(0.25)}$ & 33.22$_{(0.30)}$ & 82.91$_{(0.58)}$ & 71.12$_{(0.42)}$ & 19.75$_{(0.32)}$ & 38.97$_{(0.34)}$ & 41.11$_{(0.38)}$ & 85.08$_{(0.60)}$ & 71.55$_{(0.47)}$ \\
        \midrule
        RETAIN & No & 18.37$_{(0.28)}$ & 32.12$_{(0.38)}$ & 32.54$_{(0.27)}$ & 83.21$_{(0.43)}$ & 71.32$_{(0.32)}$ & 23.11$_{(0.47)}$ & 37.32$_{(0.36)}$ & 40.15$_{(0.41)}$ & 84.14$_{(0.34)}$ & 71.23$_{(0.38)}$ \\
        RETAIN & \textsubscript{+MPLite} & 20.42$_{(0.35)}$ & 34.56$_{(0.42)}$ & 36.87$_{(0.39)}$ & 84.73$_{(0.52)}$ & 72.94$_{(0.39)}$ & 24.85$_{(0.41)}$ & 39.68$_{(0.35)}$ & 42.67$_{(0.44)}$ & 85.82$_{(0.51)}$ & 72.83$_{(0.47)}$ \\
        \midrule
        Timeline & No & 20.46$_{(0.39)}$ & 30.73$_{(0.31)}$ & 34.83$_{(0.28)}$ & 82.34$_{(0.38)}$ & 71.03$_{(0.44)}$ & 23.76$_{(0.35)}$ & 37.89$_{(0.40)}$ & 40.87$_{(0.34)}$ & 83.45$_{(0.37)}$ & 72.30$_{(0.39)}$ \\
        Timeline & \textsubscript{+MPLite} & 22.64$_{(0.30)}$ & 32.89$_{(0.29)}$ & 36.94$_{(0.38)}$ & 83.92$_{(0.49)}$ & 72.98$_{(0.36)}$ & 24.38$_{(0.33)}$ & 39.72$_{(0.36)}$ & 42.84$_{(0.40)}$ & 84.98$_{(0.50)}$ & 73.54$_{(0.33)}$ \\
        \midrule
        GRAM & No & 20.78$_{(0.19)}$ & 34.17$_{(0.21)}$ & 35.46$_{(0.20)}$ & 81.55$_{(0.44)}$ & 68.78$_{(0.46)}$ & 24.39$_{(0.34)}$ & 38.42$_{(0.33)}$ & 41.62$_{(0.31)}$ & 85.55$_{(0.40)}$ & 69.82$_{(0.48)}$ \\
        GRAM & \textsubscript{+MPLite} & 22.78$_{(0.32)}$ & 35.96$_{(0.35)}$ & 38.61$_{(0.32)}$ & 83.22$_{(0.54)}$ & 70.94$_{(0.38)}$ & 25.93$_{(0.31)}$ & 40.42$_{(0.34)}$ & 43.68$_{(0.37)}$ & 86.98$_{(0.55)}$ & 71.06$_{(0.52)}$ \\
        \midrule
        KAME & No & 21.10$_{(0.20)}$ & 29.97$_{(0.23)}$ & 33.99$_{(0.25)}$ & 82.88$_{(0.46)}$ & 72.03$_{(0.42)}$ & 25.01$_{(0.29)}$ & 38.86$_{(0.28)}$ & 42.12$_{(0.30)}$ & 84.80$_{(0.35)}$ & 72.34$_{(0.43)}$ \\
        KAME & \textsubscript{+MPLite} & 23.64$_{(0.37)}$ & 31.98$_{(0.32)}$ & 35.92$_{(0.40)}$ & 83.67$_{(0.47)}$ & 73.48$_{(0.31)}$ & 26.22$_{(0.33)}$ & 40.74$_{(0.36)}$ & 43.95$_{(0.39)}$ & 85.92$_{(0.52)}$ & 73.95$_{(0.40)}$ \\
        \midrule
        CGL & No & 22.63$_{(0.29)}$ & 33.64$_{(0.33)}$ & 37.87$_{(0.27)}$ & 84.19$_{(0.34)}$ & 71.77$_{(0.41)}$ & 25.74$_{(0.32)}$ & 39.23$_{(0.37)}$ & 42.67$_{(0.36)}$ & 87.91$_{(0.44)}$ & 70.71$_{(0.35)}$ \\
        CGL & \textsubscript{+MPLite} & 24.82$_{(0.40)}$ & 35.68$_{(0.28)}$ & 39.97$_{(0.35)}$ & 85.82$_{(0.43)}$ & 72.81$_{(0.36)}$ & 26.97$_{(0.38)}$ & 41.92$_{(0.39)}$ & 44.61$_{(0.41)}$ & 88.89$_{(0.45)}$ & 72.52$_{(0.39)}$ \\
        \midrule
        G-BERT & No & 22.28$_{(0.25)}$ & 35.62$_{(0.29)}$ & 36.46$_{(0.26)}$ & 81.50$_{(0.38)}$ & 71.18$_{(0.43)}$ & 25.12$_{(0.30)}$ & 39.91$_{(0.31)}$ & 43.25$_{(0.28)}$ & 85.76$_{(0.50)}$ & 72.88$_{(0.45)}$ \\
        G-BERT & \textsubscript{+MPLite} & 24.31$_{(0.36)}$ & 37.14$_{(0.30)}$ & 38.98$_{(0.33)}$ & 82.99$_{(0.56)}$ & 72.72$_{(0.42)}$ & 27.58$_{(0.35)}$ & 42.56$_{(0.34)}$ & 44.62$_{(0.39)}$ & 86.88$_{(0.58)}$ & 74.12$_{(0.47)}$ \\
        \midrule
        HiTANet & No & 23.15$_{(0.28)}$ & 34.68$_{(0.35)}$ & 35.97$_{(0.31)}$ & 85.13$_{(0.31)}$ & 73.15$_{(0.39)}$ & 24.53$_{(0.33)}$ & 38.42$_{(0.37)}$ & 41.89$_{(0.29)}$ & 86.34$_{(0.36)}$ & 71.35$_{(0.44)}$ \\
        HiTANet & \textsubscript{+MPLite} & 25.87$_{(0.33)}$ & 36.91$_{(0.36)}$ & 39.02$_{(0.34)}$ & 86.74$_{(0.47)}$ & 74.45$_{(0.40)}$ & 26.91$_{(0.30)}$ & 42.12$_{(0.38)}$ & 43.94$_{(0.33)}$ & 87.99$_{(0.49)}$ & 72.85$_{(0.42)}$ \\
        \bottomrule
    \end{tabular}
\end{table*}

\subsection{Dataset Description}
\label{sec:dataset}
To evaluate our proposed model, we focus on two pubic and widely-used EHR datasets: MIMIC-III~\cite{johnson2016mimic} and MIMIC-IV~\cite{johnson2020mimic}.
Both datasets are derived from extensive de-identified clinical data collected from patients admitted to Intensive Care Units (ICUs).
We employed a randomized approach to divide both datasets into training, validation, and testing segments.
Specifically, MIMIC-III and MIMIC-IV datasets were divided into 6000/493/1000 and 8000/1000/1000 for the training, validation, and test sets, respectively.

Table \ref{tab:statistics} shows the basic statistics in MIMIC-III. 
Note that, while there are 85,155 patients in MIMIC-IV with multiple visits, we remove the patients with the overlapped time range and then randomly sample 10,000 patients from MIMIC-IV from 2013 to 2019 for training, which retains the same setting as Chet~\cite{LuH022}.
Hence, the basic statistics of MIMIC-IV which is omitted in paper might change for every runtime, since the random sampling method is adopted to get the comparable sample size of patients with MIMIC-III. 
We select patients with multiple admission records (\# of visits $\geq$ 2) for the diagnosis prediction task and only consider patients without missing diagnosis codes in all visits.
For instance, among the 38,983 single-visit patients in MIMIC-III, we only consider patients with previous lab results before their admission.
We conduct multiple experiments with uniform baseline hyperparameter sets, measuring average and standard deviation values.

\subsection{Baselines}
To check the improvement of MPLite for predictive models, we select the following state-of-art methods as baselines:

\begin{itemize}
    \item RNN/CNN-based models: GRU~\cite{ChoMGBBSB14}, Timeline~\cite{BaiZEV18}, RETAIN~\cite{ChoiBSKSS16}, Deepr~\cite{NguyenTWV17}, and Dipole~\cite{MaCZYSG17}.
    \item Graph-based models: GRAM~\cite{ChoiBSSS17}, KAME~\cite{MaYXCZG18}, and CGL~\cite{LuRCKN21}.
    \item Transformer-based models: G-BERT~\cite{ShangMXS19}, HiTANet~\cite{LuoYXM20}.
\end{itemize}

Note that GRU uses multi-hot vectors of medical codes as inputs, while other baselines use medical code embeddings.
For G-BERT, both pretraining and medication inputs is discarded which requires extra information other than diagnose features.
Moreover, we remove the clinical notes parsing module in CGL and the timestamp feature in HiTANet to ensure each baseline is trained by the same data.
We also do not consider MiME~\cite{ChoiXSS18} and GCT~\cite{ChoiXLDFXD20} because of different evaluation tasks.

\subsection{Parameters Setting}
The parameter settings used for pretraining module, we find the optimal output dimension 200 of the first dense layer from a search space of [100, 200] and set Drop-out rate as 0.4 in the final classifier for fine-tuning and final prediction.
For the baseline GRU, the units of RNN module are all set as 128.
For other baselines, we do our best to follow the parameter setting described in original papers.
Different learning rate decay schedulers with Adam optimizer are experimented, resulting a decay learning rate from \(1e-2\) to \(1e-5\) between epochs.
Moreover, we set batch size as 64 and use 100 epochs for training process.
We conduct 10 repeated experiments for each baseline model and the corresponding model with pretrained lab results. 
All evaluation metrics are recorded and calculated for each experiment, and we can then assess whether \textit{MPLite} can help model get more accurate prediction.
The source code of MPLite will be released upon publication.

All programs are implemented using Python 3.10, Tensorflow 2.10, and Pytorch 2.3.1 with CUDA 12.3 on a machine with two AMD EPYC 9254 24-Core Processors, 528GB RAM, and four Nvidia L40S GPUs.

\subsection{Evaluation Metrics}
Since among the 4880 and 6102 diseases we are predicting in MIMIC-III and MIMIC-IV, the distribution of disease codes is very sparse and the occurrence for each disease is highly imbalance, we adopt the weighted F1 score (w-F\textsubscript{1} \cite{BaiZEV18}) and top k recall (R@k \cite{ChoiBSKSS16}) for diagnosis predictions.
In the context of the weighted F\textsubscript{1} score, the contributions of individual classes (diseases) are weighted based on their prevalence in our dataset. Unlike traditional recall, Recall@k focuses on the ratio of true positive samples among positive samples in the top k predictions, and we set k values as 10 and 20 for evaluation which is the same as other works.
For heart failure predictions in case study, we add the area under the ROC curve (AUC) as binary classification metrics besides F\textsubscript{1} score, since label distribution is imbalanced in MIMIC datasets.

\subsection{Experimental results}
\subsubsection{Diagnosis Prediction}

As demonstrated in Table \ref{tab:mimic_combined}, both the mean and standard deviation are reported across different baselines within two datasets. The results indicate that the integration of the proposed framework consistently enhances the predictive performance of various baselines.

From the results on the MIMIC-III dataset, we observe significant improvements in w-F\textsubscript{1} scores when the proposed framework is applied. For example, GRU with \textit{MPLite} improves over the vanilla GRU by approximately +1.76 in w-F\textsubscript{1}, +2.26 in R@10, and +2.33 in R@20. This trend is consistent across other models such as Dipole, Deepr, and RETAIN, demonstrating similar enhancements in w-F\textsubscript{1} and recall metrics. Specifically, Dipole with \textit{MPLite} achieves a w-F\textsubscript{1} score improvement of +3.61, and an increase of +2.18 in R@10 and +3.53 in R@20, highlighting the efficacy of pretraining with \textit{MPLite}.

On the MIMIC-IV dataset, the improvement trends are similar. GRU shows an increase of +1.32 in w-F\textsubscript{1} and notable gains of +2.72 in R@10 and +2.72 in R@20 with the pretraining module. These results suggest that \textit{MPLite} not only boosts w-F\textsubscript{1} scores but also enhances recall rates, indicating better model sensitivity in capturing relevant diagnostic information.

The consistent performance boost across both datasets underscores the generalization capability of the proposed module. 
The most likely reason for the improved performance is that lab results typically include detailed physiological and biochemical indicators, which directly reflect patients' health status, providing crucial information about disease conditions and bodily functions for doctors to diagnose diseases.

\subsubsection{Case Study - Heart Failure Prediction}
Following the results of diagnosis predcition, a research question arises: \textit{Assuming different training tasks, where the final prediction and the proxy task in the pretraining module are different, can MPLite still improve the predictive performance of the baseline models?}
Table \ref{tab:mimic_combined} also shows the heart failure prediction results in both MIMIC-III and MIMIC-IV.
We observe that \textit{MPLite} still allows all involved baselines to achieve higher AUC and F\textsubscript{1} scores.
Therefore, we conclude that lab results can complement other clinical information, contributing collectively to precise prediction, which also demonstrates the generalization ability of the proposed framework.

\section{Discussion and Conclusion}
In this paper, we present a flexible plug-in-and-play framework called \system{}, which integrates lab results to enable backbone models to collaboratively learn more precise representations for patients.
We conduct experiments on 2 widely-used EHR datasets across 10 predictive baselines with different architectures, demonstrating the effectiveness of the plug-in pretraining module through significant improvements over the original backbone models.
Additionally, we performed a case study on heart failure prediction to verify the generalization ability of \textit{MPLite} across various prediction tasks.
All pre-training experiments are based on lab results which can be limited when such features are not available.
While the framework can be adapted to other input types, we believe further extensive testing is necessary.

In the future, we plan to evaluate the effectiveness of the pretraining model on more complex network architectures than MLP and diverse health risk prediction tasks.
For instance, we can further refine proxy tasks to help model get rid of limitation on laboratory input, which is also the common problem as other baselines upon lab tests.
Furthermore, an initial screening process could be applied to single-visit patients to enhance training quality by ensuring adequate diversity of single-visit and multi-visit patients.
Another potential direction for future research is to incorporate more feature modalities such as clinical notes into the pre-training process.

\bibliographystyle{IEEEtran}
\bibliography{main}


\end{document}